# Narrative as a Dynamical System

Doxas[1,2], I., J. Meiss[3], S. Bottone[1], T. Strelich[4,5], A. Plummer[5,6], A. Breland[5,7], S. Dennis[8,9], K. Garvin-Doxas[9,10], M. Klymkowsky[3]

[1]*Northrop Grumman Corporation*
[2]*Some work performed at the University of Colorado, Boulder*
[3]*University of Colorado, Boulder*
[4]*Fusion Constructive LLC*
[5]*Work performed at Northop Grumman Corporation*
[6]*Current Address JP Morgan*
[7]*Current address, GALT Aerospace*
[8]*University of Melbourne*
[9]*Work performed at the University of Colorado, Boulder*
[10]*Boulder Internet Technologies*

There is increasing evidence that human activity in general, and narrative in particular, can be treated as a dynamical system in the physics sense; a system whose evolution is described by an action integral, such that the average of all possible paths from point A to point B is given by the extremum of the action. We create by construction three such paths by averaging about 500 different narratives, and we show that the average path is consistent with an action principle.

## Introduction

In her recent article "The Geometry of Thought", outgoing Association for Psychological Science president Barbara Tversky (1) described humans' "spatial thinking" and our tendency to embed objects and experiences in a space that we can navigate for recollection and sense-making. This picture is largely in agreement with the mathematical and computational apparatus often built for Machine Learning (ML), where objects and actions are embedded in high-dimensional spaces, with one significant addition: for ML applications we embed objects in *metric* spaces. Then the question becomes: how do we navigate; how do we "move" in a metric space?

The present work rests on the emerging evidence that human activity in general, and narrative description in particular, is a dynamical system in the physics sense; a system whose evolution is governed by an Action Integral

$$I = \int_A^B L\, ds \qquad [1]$$

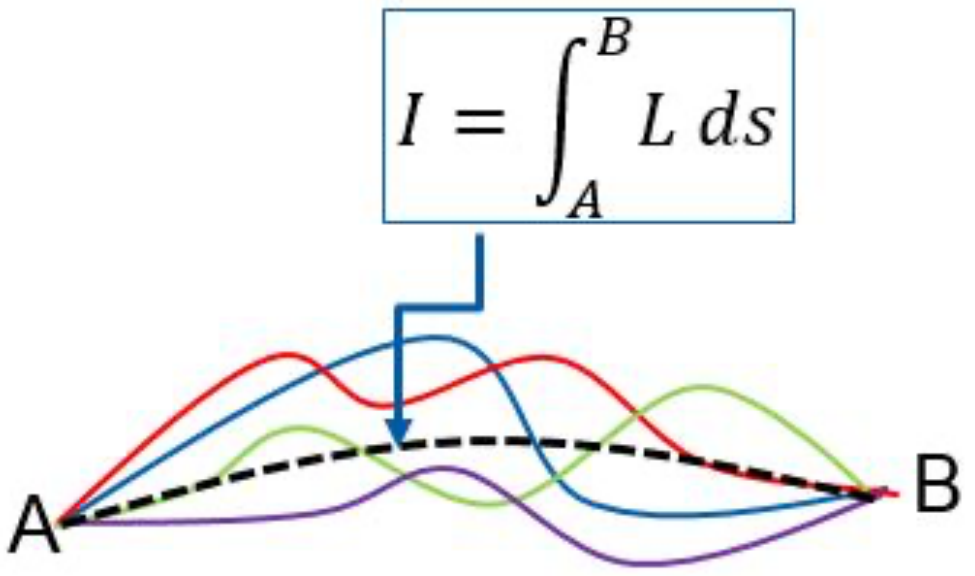

**Figure 1.** Of all the paths that connect A to B, a dynamical system follows the path for which the action, $I$, is an extremum.

where $L$ is the Lagrangian of the system and $s$ is some parameter along the path, usually time. The evolution of such a system from point A to point B proceeds along the Classical Path (cf. Figure 1), which is the path for which the Action

Integral (Eq.-1) is an extremum. Well-known principles like the Fermat Principle (for ray tracing) and the Hamilton Principle (for classical Mechanics) are expressed as action integral formulations. The degree to which "human activity" can be described by a dynamical system is an emerging area of research, but there is a steady pace of new examples and analyses over recent years that support the underlying hypothesis (e.g. 4, 13, 16).

*Evidence of Dynamics*

Both the evidence for the dynamical provenance of the data, and some of the methods we employ, rest on the Takens theorem (2), which states that the delay embedding of almost any observable of a dynamical system has the same topological properties as the system.

Appreciating the significance of this theorem requires an understanding of the delay embedding technique. Simply stated, this technique transforms a timeseries into a high-dimensional object by designating each successive value in the timeseries as the coordinate for a different dimension.

To illustrate the method, consider a timeseries of the X-coordinate of the Lorenz system (3), $[X(t), X(t-\tau), X(t-2\tau), \ldots]$, and a function (observable) S(X), so that the timeseries of the observable S is given by

$$[S(X(t)),\ S(X(t-\tau)),\ S(X(t-2\tau)), \ldots] \equiv [S(t), S(t-\tau), S(t-2\tau), \ldots] \qquad [2]$$

where $\tau$ is the delay. The points of the delay embedding of the timeseries S into an N-dimensional embedding space are then

$$[\{S(t), S(t-\tau), \ldots, S(t-(N-1)\tau\}, \{S(t-\tau), S(t-2\tau), \ldots, S(t-N\tau)\}, \ldots][3]$$

so that the coordinates of each point that is embedded in the N-dimensional space are N successive values of the timeseries. Figure-2 illustrates this concept on two separate embeddings, $M_X$ and $M_Y$, for two of the coordinates of the Lorenz system (3). Note that Figure-2 replaces the observable $S(X)$ with $X$ for simplicity, but this does not change the result; in principle any function of $X$ will yield the same result.

It is easy to miss the power of the theorem and spend a lot of effort trying to find "relevant" observables. "Any observable" of a dynamical system really does mean almost *any* observable – the theorem only requires a very generic function (2). For example, the theorem has been applied to several different observables of the solar wind: temperature, speed, northward magnetic field, density, etc. It has also been successfully applied to composite quantities like the Amplitude Lower (AL) index, which is a complicated average of magnetometer readings around the northern hemisphere. To within the fidelity of each set of measurements, they all give the same answer.

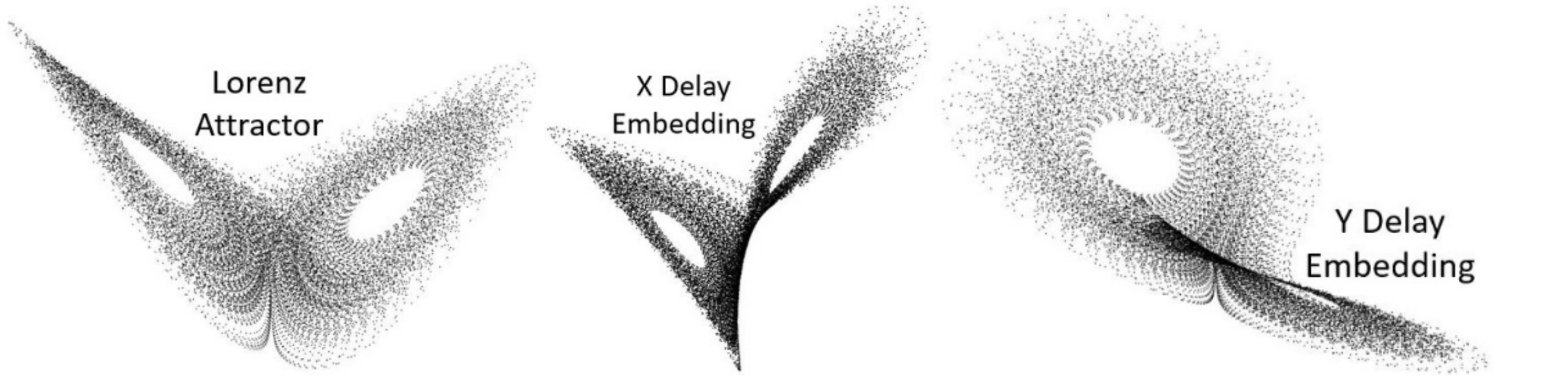

**Figure 2.** The full Lorenz attractor and two delay embeddings using the X and Y coordinates, respectively.

As previously mentioned, human behavior can also be described as a dynamical system in the sense that decisions made over time cause one state to evolve into another. This argument is supported by prior research demonstrating that social systems also exhibit the Takens property. For example, dynamical analysis of lifelog images – images taken at various intervals of 3-10 min over the period of several days by body cameras – show that the images people see as they move about their lives are a dynamical system (4). Moreover, even obtuse observables like the timeseries of the first singular value of the images were shown to share the same properties as the overall system. The same argument applies to a surprisingly diverse set of domains, including natural language. Reference (5) for example describes the dynamics involved in the way language changes over generations, and (6) describes the dynamics of the emergence of grammatical rules within a single mind. Narrative text - from *War and Peace* (see Figure-3), to news articles, to narrative descriptions of auto parts – can also be treated as dynamical systems because different observables have been shown to share the same embedding properties. These examples reflect the diversity of the applications of dynamical systems research, but are by no means exhaustive.

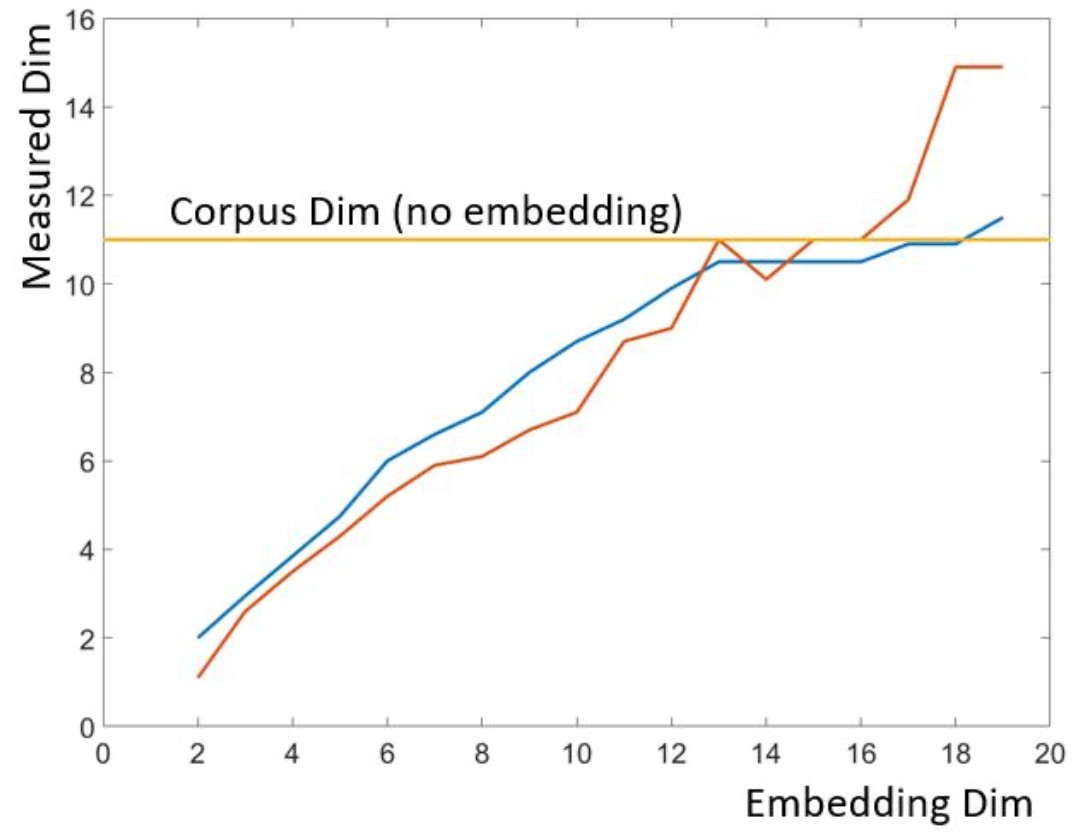

**Figure 3.** The measured dimension of the delay embedding of the first singular value of paragraphs in War and Peace as a function of the embedding dimension. The blue curve uses the delay embedding of the text in the correct sequence, the red curve uses the delay embedding of the text paragraphs in random order. The horizontal line is the dimension of the corpus (no delay-embedding). We see that the dimensionality of delay-embedding of the correctly ordered paragraphs tends to that of the corpus, while the dimensionality of the shuffled paragraphs continues to increase. The result is similar to that of (4) for lifelog images and suggests that the paragraphs of War and Peace are an observable of a dynamical system.

## Results

*Constructing a Classical Path*

The classical path (cf. Figure 1) is the mean of all paths that the system can trace from A to B. To further explore this formulation of the evolution of narrative, we constructed an ensemble of narratives in such a way as to have a readily computable mean (Figure 4). We use three groups of narratives which all have the same number of paragraphs, allowing by construction for the ready computation of the mean for each paragraph. By fixing the first and one other paragraph in each group, we anchor the end points of the action integral for all narratives in the group. Each paragraph is at least 40 words long so that we can meaningfully use a number of different semantic embedding methods.

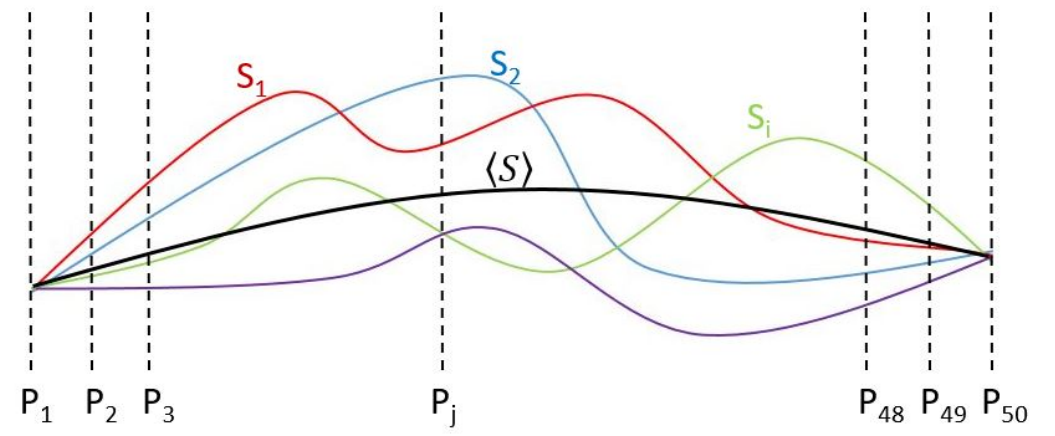

**Figure 4.** The construction of the classical path. All narratives have the same number of paragraphs so that the average path can be easily computed.

Each group is composed of the stories submitted to a different literary contest, one of the Owl Canyon Press Short Story Hackathons (7-9). For each contest, the authors were given two paragraphs, and produced the rest. Other than the total number of paragraphs, and the requirement that each paragraph have at least 40 words, no other restrictions were placed on the stories; each group contains stories that cover a wide range of genres. The top rated stories for each Hackathon have been published in anthologies by Owl Canyon Press (7-9), but the analysis presented here used all stories that satisfied the criteria of number and length of paragraphs.

Figure 5 shows the structure of each of the three narrative groups, and the number of narratives, $N$, in each. The narratives in Group-1 (G1) and Group-2 (G2) all have 50 paragraphs. For G1 the first and last paragraphs are the same for all narratives, and for G2 the first and the $20^{th}$ paragraphs are the same for all narratives. Narratives in Group-3 (G3) have a total of only 20 paragraphs, with the first and last being common to all narratives.

We use two different methods to embed the paragraphs of the narratives into a high-dimensional metric space: Latent Semantic Analysis (LSA, e.g. 10), and doc2vec (11, 12). The results of these two methods are very similar, a finding that is consistent with previous work (e.g. 13), which used three different embedding methods, all giving similar results (LSA; Non-negative Matrix Factorization, (14); and the Topics Model, (15)). For both LSA and doc2vec we use 300 embedding dimensions, but we tested variability with dimensionality using between 100 and 500 dimensions, with no appreciable difference in the results, also consistent with previous work (13, 16).

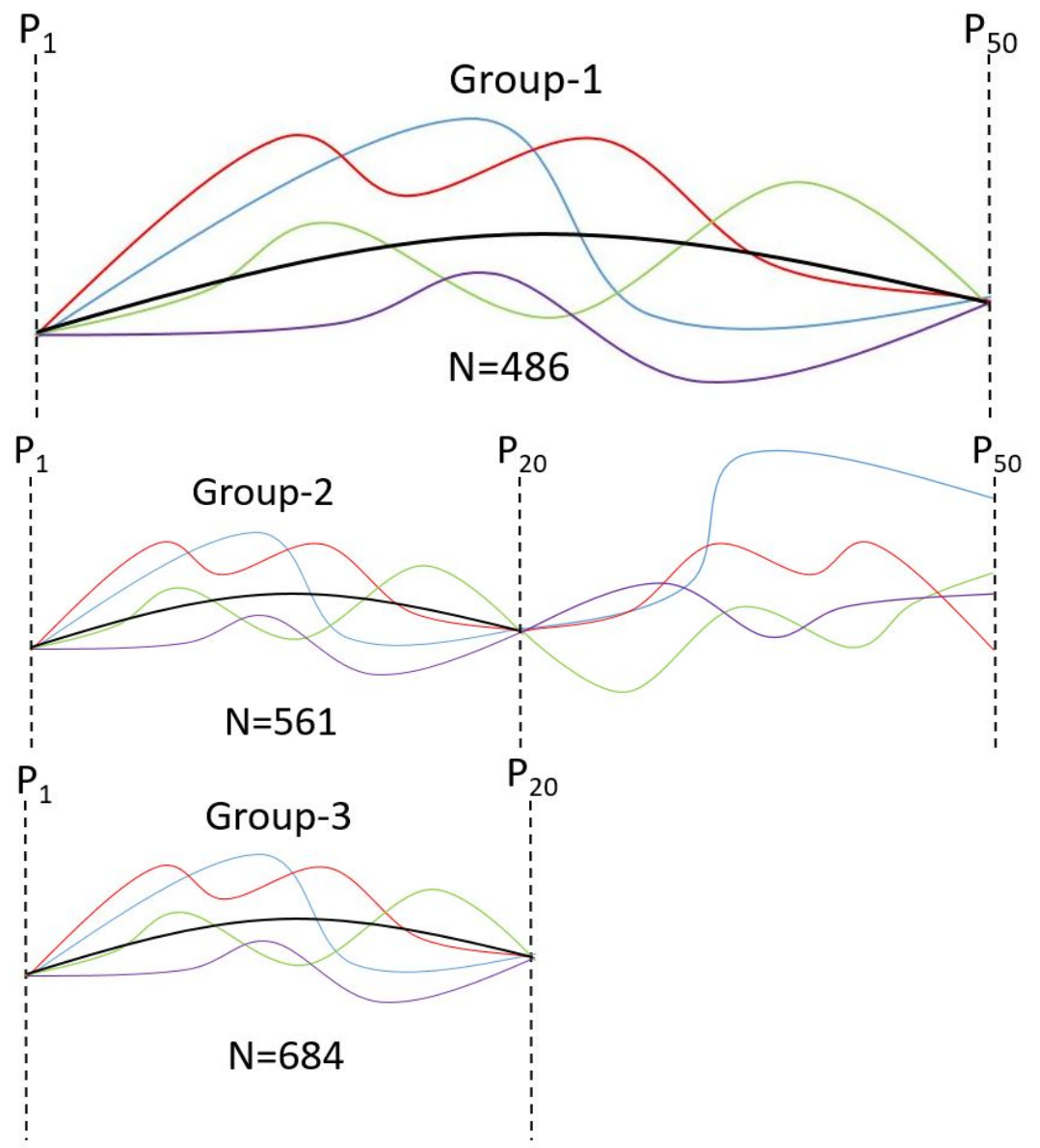

**Figure 5.** The path structure of the narrative groups. All narratives in each group have the same number of paragraphs so that the average path can be easily computed, and each group has two anchor paragraphs in common. For Group-1 these are $P_1$ and $P_{50}$, and for the other two these are $P_1$ and $P_{20}$. N is the number of narratives in each group.

If we denote by $P_{ij}$ the jth paragraph of the ith narrative, $1 \leq i \leq N, 1 \leq j \leq n$, the average paragraphs for each group are given by

$$\langle \vec{P}_j \rangle = \frac{1}{N} \sum_{i=1}^{N} \vec{P}_{ij} \quad j = 1, \dots, n \quad [4]$$

so that the jth average paragraph is the average of the jth paragraphs of the $N$ narratives in the group. The average narrative, $\langle S \rangle$, is then given by the series

$$\{\langle \vec{P}_1 \rangle, \langle \vec{P}_2 \rangle, \langle \vec{P}_3 \rangle, \dots, \langle \vec{P}_n \rangle\}$$

where $n$ is the number of paragraphs in each narrative, $1 \leq j \leq n$.

Constructing the Lagrangian in the Action Integral is not an obvious exercise for non-physical systems like narratives. Nevertheless, we take a formal approach, we define the Lagrangian as

$$L = T - V \quad [5]$$

and justify the approach *a posteriori*. In Eq. [5] $T = (1/2)mv^2$ is the kinetic energy and $V$ is the potential energy. We again take a formal approach, write $\vec{v} = \Delta\vec{P}/\Delta t$, and consider potential-free motion, so that $V = 0$. With these definitions, the Action becomes

$$I = \frac{m}{2} \sum_{\substack{\text{path} \\ j=2}}^{n} \frac{\left(\langle \vec{P}_j \rangle - \langle \vec{P}_{j-1} \rangle\right)^2}{(\Delta t)^2}$$

$$= \alpha \sum_{\substack{\text{path} \\ j=2}}^{n} \left(\langle \vec{P}_j \rangle - \langle \vec{P}_{j-1} \rangle\right)^2 \quad [6]$$

where we discretized the integral into a sum over the $n$ paragraphs and collected all constant factors into the term $\alpha$. We let the narrative time be one step per paragraph by taking $t_i = i$, so that $\Delta t = 1$ is a constant. Since $\alpha$ is a constant, minimizing the Action is equivalent to minimizing the sum of the square distances between successive paragraphs along the path.

To investigate the proposition that the average path corresponds to the minimum of the Action Integral, we compute for each average path the Minimum Spanning Tree (MST), and the Traveling Salesman Problem (TSP), the latter modified to allow for different starting and ending points. The TSP asks "which sequence of points gives the lowest action along the path" and is therefore a computational expression of the Action Principle. The MST asks "which configuration of points gives the lowest sum", which allows for points to lie outside the direct path from A to B and is therefore more geometric than dynamical. Nevertheless, by showing that even when considering the more general configuration, the minimum sum for the average narrative still contains long correctly-ordered linear stretches, it can provide better corroboration for this paper's main thesis, that narrative can be modeled as a dynamical system.

Figures 6(a), 7(a), and 8(a) show the MSTs for the three narrative groups. We see that, even though the MST does not require all points to lie on a line, the minimum sum configuration includes in all cases an almost linear run of the first 10 paragraphs that are in approximately the correct order at the start of the average narrative. Furthermore, we see that for the groups that end at the second anchor point (G1 and G3) the last eight paragraphs also form an almost linear group in the correct order. For narratives that are longer than 20 paragraphs, we see that the MSTs become less linear and more unordered in the middle. It is worth noting that for G2, which continues after the second anchor paragraph, there

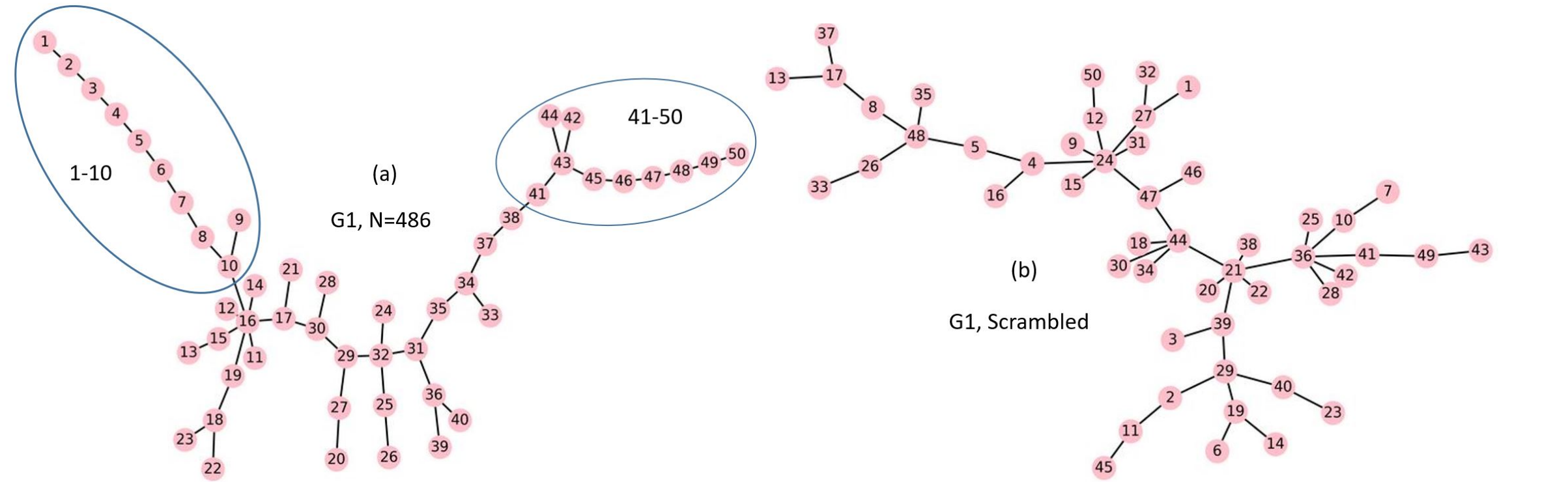

**Figure 6**: The Minimum Spanning Trees (MSTs) for the average narrative for Group-1. (a) The paragraphs are in the original storytelling order. (b) The order of the paragraphs in each story has been randomly scrambled as described by Eq. 7.

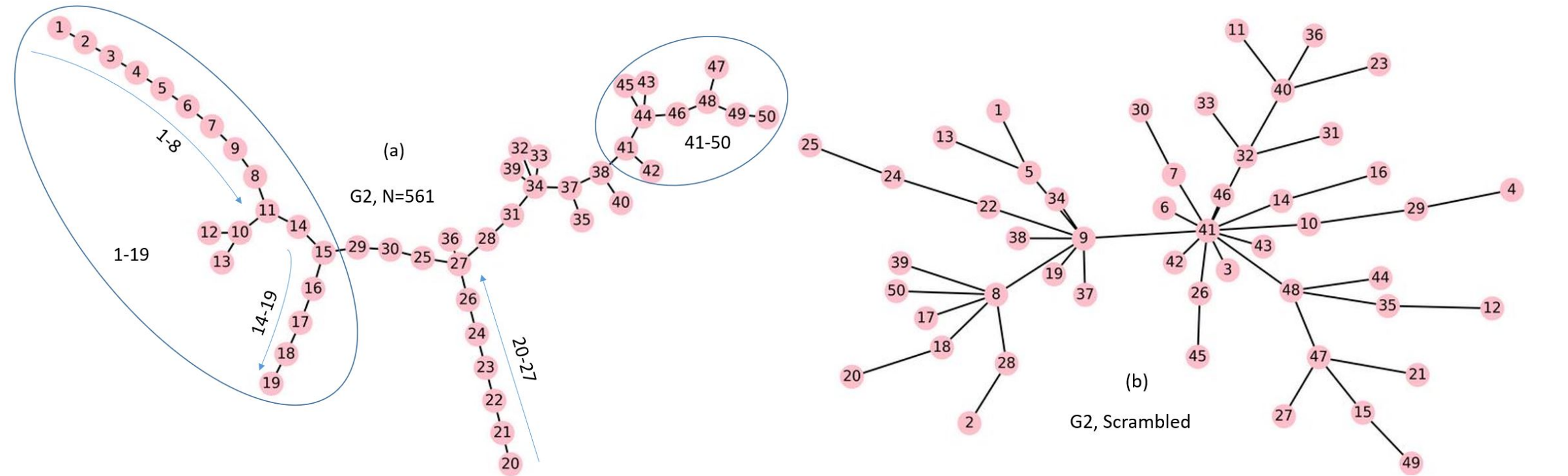

**Figure 7**: The Minimum Spanning Trees (MSTs) for the average narrative of Group-2. (a) The paragraphs are in the original storytelling order. (b) The order of the paragraphs in each story has been randomly scrambled as described by Eq. 7.

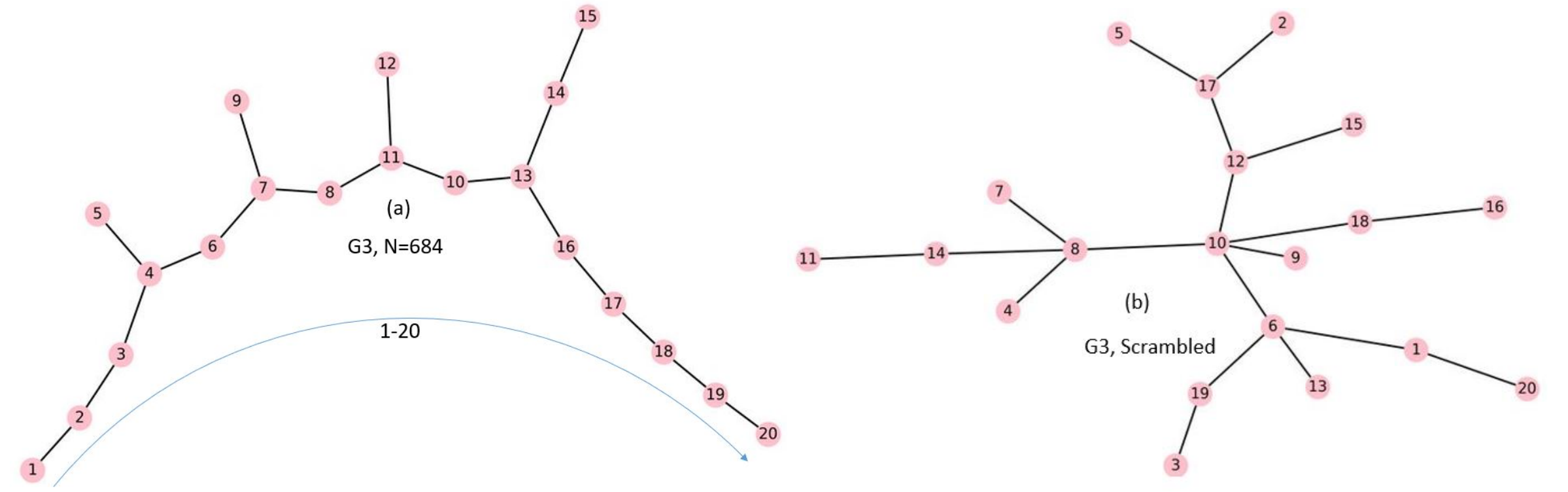

**Figure 8**: The Minimum Spanning Trees (MSTs) for the average narrative of Group-3. (a) The paragraphs are in the original storytelling order. (b) The order of the paragraphs in each story has been randomly scrambled as described by Eq. 7.

is an almost linear run following that anchor as well (paragraphs 20-27), although the anchor paragraph appears at the end of a graph branch, as if it were a new start. In Figure 9 we split these branches, emphasizing the near-correct ordering of each part.

To check that the observed MST structure is the result of the proper ordering of the paragraphs in the narratives (i.e. of the dynamics of narrative) and not a property of the text as a whole, we repeat the MST calculations after shuffling the paragraph order in each narrative In this case each narrative in a group has the same paragraphs as before, but now they are in a random order instead of the correct story-telling order. This means that $\langle \vec{P}_j \rangle$ is now

$$\langle \vec{P}_j \rangle = \frac{1}{N} \sum_{i=1}^{N} \vec{P}_{i\pi(i)_j} \qquad [7]$$

where $\pi(i)_j$ is the $j$th component of a random permutation of the $n$ paragraphs of narrative $i = 1, \ldots, N$. Each $\langle \vec{P}_j \rangle$ is therefore still the average of $N$ paragraphs, one from each narrative, but no longer the jth one from each narrative. Figures 6(b), 7(b) and 8(b) show the results. We see that the almost-linear structures at the beginning and end of the average narrative have disappeared and that the average paragraphs no longer appear in the correct order. We have paired each shuffled-paragraph figure with the correctly-ordered one for each group for ease of comparison.

Table-1 shows the results of the modified Traveling Salesman Problem (TSP) for the three groups of narratives. We see that when we restrict the computation of the sum to the path, as in [6], and we take a strict definition of the path to only extend between the anchor points (A and B in Figure-1), the results are similar to those of the MST, with correctly ordered runs at beginning and end. When we repeat the calculations with the shuffled paragraph order, the correctly ordered runs no longer appear. The table also

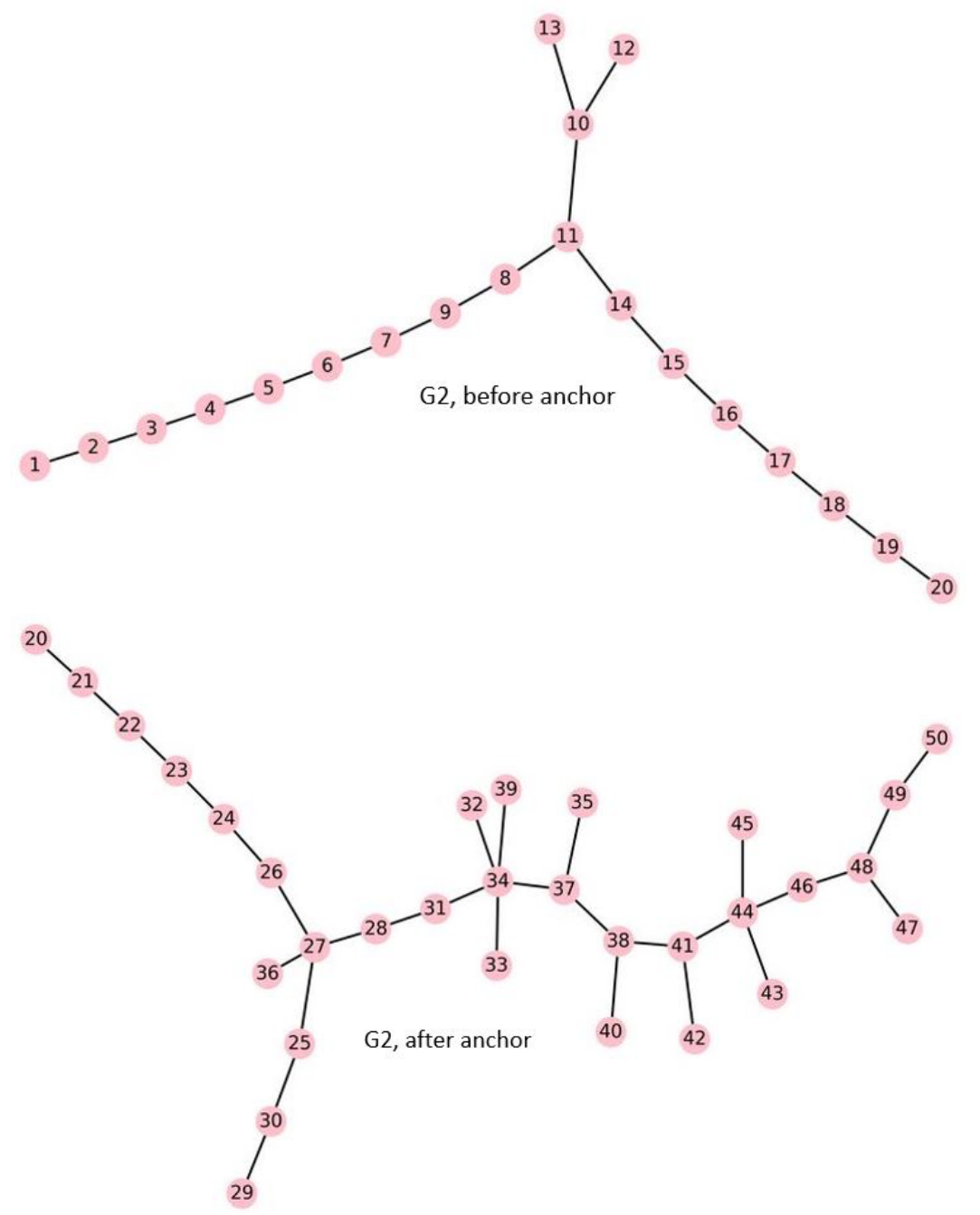

**Figure 9**: The MSTs for the two brances of Group-2 on either side of the anchor paragraph-20. The MSTs are the same as in Figure 8, but separating them makes clear the linear, correctly ordered runs before and after the middle anchor paragraph.

pairs the shuffled with the unshuffled results for each narrative group for ease of comparison.

## Conclusions

As a narrative evolves, be it a description or a story-telling, successive paragraphs can be considered as tracing a trajectory in some high-dimensional semantic space. It is of course obvious that not any such succession of points will correspond to a comprehensible narrative, and the question then becomes, what laws, if any, govern the evolution of a narrative. It has been shown (13) that when considered as a "bag of paragraphs", with no consideration to order, narratives occupy a surprisingly low-dimensional space (at least compared with the number of embedding dimensions often used). What we show above is that the evolution is consistent with a dynamical description in the physics sense, in which narratives evolve according to an Action Principle: the average narrative is the one that connects the anchor paragraphs with the minimal action, Eq.-[1]. The TSP results show that the lowest-Action path is very close to the average narrative, at least near the anchor points, and the MST results show that even when considering the more general problem of the minimum-sum tree structure, the result is still very similar to the average path near the anchor points.

| | | | | | | | | | | | | | | | | | | | | | | | | |
|---|---|---|---|---|---|---|---|---|---|---|---|---|---|---|---|---|---|---|---|---|---|---|---|---|
| Group-1, N=486 | | | | | | | | | | | | | | | | | | | | | | | | |
| 1 | 2 | 3 | 4 | 5 | 6 | 7 | 8 | 9 | 10 | 11 | 13 | 15 | 14 | 16 | 12 | 17 | 21 | 26 | 25 | 28 | 30 | 29 | 27 | 20 |
| 19 | 18 | 22 | 23 | 24 | 32 | 33 | 34 | 37 | 35 | 31 | 36 | 39 | 40 | 38 | 41 | 42 | 43 | 44 | 45 | 46 | 47 | 48 | 49 | 50 |
| Shuffled | | | | | | | | | | | | | | | | | | | | | | | | |
| 1 | 28 | 2 | 38 | 29 | 20 | 4 | 27 | 32 | 30 | 19 | 49 | 16 | 39 | 37 | 31 | 33 | 46 | 42 | 17 | 7 | 13 | 5 | 47 | 22 |
| 45 | 14 | 34 | 6 | 44 | 9 | 36 | 12 | 24 | 48 | 26 | 10 | 35 | 40 | 21 | 15 | 25 | 41 | 43 | 23 | 3 | 11 | 18 | 8 | 50 |
| Group-2, N=561 | | | | | | | | | | | | | | | | | | | | | | | | |
| 1 | 2 | 3 | 4 | 5 | 6 | 7 | 9 | 8 | 11 | 10 | 13 | 12 | 14 | 15 | 16 | 17 | 18 | 19 | 20 | 21 | 22 | 23 | 24 | 26 |
| 25 | 30 | 29 | 32 | 36 | 27 | 28 | 31 | 33 | 34 | 39 | 40 | 38 | 37 | 35 | 42 | 41 | 43 | 44 | 45 | 46 | 47 | 48 | 49 | 50 |
| Shuffled | | | | | | | | | | | | | | | | | | | | | | | | |
| 1 | 17 | 14 | 13 | 9 | 7 | 2 | 6 | 3 | 12 | 18 | 15 | 5 | 10 | 11 | 4 | 19 | 16 | 8 | 20 | 43 | 32 | 38 | 28 | 29 |
| 48 | 39 | 35 | 23 | 42 | 34 | 47 | 31 | 27 | 36 | 33 | 25 | 30 | 21 | 26 | 40 | 37 | 22 | 24 | 44 | 49 | 46 | 45 | 41 | 50 |
| Group-3, N=684 | | | | | | | | | | | | | | | | | | | | | | | | |
| 1 | 2 | 3 | 4 | 5 | 6 | 8 | 7 | 9 | 10 | 11 | 12 | 13 | 14 | 15 | 16 | 17 | 18 | 19 | 20 | | | | | |
| Shuffled | | | | | | | | | | | | | | | | | | | | | | | | |
| 1 | 9 | 13 | 18 | 15 | 6 | 2 | 7 | 8 | 12 | 10 | 19 | 5 | 4 | 16 | 3 | 17 | 11 | 14 | 20 | | | | | |

**Table 1**. The order of the average paragraphs for each narrative group as given by the Traveling Salesman Problem (TSP) modified to allow for different initial and final points. We see that they all have long correctly ordered runs at beginning and end (highlighted in blue). Group-3 in particular is almost perfectly ordered, with only one inversion (7↔8).

There is one additional observation we can make about the results above. We see that in both the MST and the TSP analyses, the ordered linear runs appear at the beginning and end of the average narrative, after and before the anchor points, while the intermediate paragraphs appear to be more disordered. This is particularly clear for G2, where the second anchor paragraph is in the middle of the narrative, instead of the end (recall the separated MST plots in Figure 9). We see that the linear, correctly ordered runs appear before and after the anchor point at paragraph-20. Moreover, notice that, even though the ends of the individual narratives in the group are not fixed, the average still has a well define linear run, in the correct order, at the end of the story (Table-1). Further investigation is needed to ascertain whether this is a more general property that relates to the exigencies of storytelling.

## Acknowledgments

ID and SD wish to acknowledge many stimulating discussions with the late prof. Walter Kintsch. This work was partly supported by a grant from the National Science Foundation.

## References

1. B. Tversky. What's on my mind: The Geometry of Thought. Available at [https://www.edge.org/ conver-sation/barbara_tversky-the-geometry-of-thought (2019)](https://www.edge.org/conversation/barbara_tversky-the-geometry-of-thought)
2. F. Takens. Detecting strange attractors in turbulence. In D. A. Rand & L. S. Young (Eds.), *Lecture notes in mathematics* (pp. 366–381). Berlin, Germany: Springer-Verlag (1981).
3. E. Lorenz. Deterministic nonperiodic flow. *Journal of the Atmospheric Sciences*. **20** (2): 130–141 (1963).
4. V. Sreekumar, S. Dennis, I. Doxas, Y. Zhuang, & M. Belkin, M. The geometry and dynamics of lifelogs: Discovering the organizational principles of human experience. *PLoS ONE*, 9(5), e97166. doi:10.1371/journal.pone.0097166 (2014).
5. P. Niyogi and R. C. Berwick. A Dynamical Systems Model for Language Change, *Complex Systems*, 11, 161–204 (1997).
6. J. L. Elman. Language as a dynamical system. In R. F. Port & T. van Gelder (Eds.), *Mind as motion: Explorations in the dynamics of cognition* (pp. 195–225). The MIT Press (1995).
7. Owl Canyon Press. No bars and a dead battery. The Summer 2018 Owl Canyon Press Short Story Hackathon Contest Winners, Gene Hayworth Ed., *Owl Canyon Press*, Boulder, CO (2018).
8. Owl Canyon Press. When the ride ends. The Autumn 2019 Owl Canyon Press Hackathon winners, Gene Hayworth Ed., *Owl Canyon Press*, Boulder, CO (2020).
9. Owl Canyon Press. An odd sized casket. The Autumn 2020 Owl Canyon Press Short Story Hackathon Contest winners, Gene Hayworth Ed., *Owl Canyon Press*, Boulder, CO (2021).
10. T. K. Landauer, P. W. Foltz & D. Laham, D. An introduction to latent semantic analysis. *Discourse Processes*, 25(2–3), 259–284 (1998).
11. T. Mikolov, K. Chen, G. Corrado, and J. Dean. Efficient Estimation of Word Representations in Vector Space. arXiv:1301.3781v3 [cs.CL] (2013a).
12. T. Mikolov, I. Sutskever, K. Chen, G. Corrado, and J. Dean (2013b). Distributed Representations of Words and Phrases and their Compositionality. arXiv:1310.4546v1 [cs.CL] (2013b).
13. I. Doxas, S. Dennis, and W. L. Oliver. The dimensionality of discourse. *Proc. Natl. Acad. Sci. U.S.A.*, 107(11), 4866–4871 (2010).
14. D. Lee and H. Seung. Algorithms for non-negative matrix factorization. *Adv Neural Inf Process Syst* 13:556–562 (2001).
15. D. Blei, T. Griffiths, M. Jordan, and J. Tenenbaum. Hierarchical topic models and the nested Chinese restaurant process. *Adv Neural Inf Process Syst* 16:17–24 (2004).
16. V. Sreekumar, S. Dennis, and I. Doxas. The Episodic Nature of Experience: A Dynamical Systems Analysis. *Cognitive Science*, DOI: 10.1111/cogs.12399 (2016).